\documentclass[conference]{IEEEtran}
\pdfoutput=1
\usepackage{cite,amsmath,amssymb,amsfonts}
\usepackage{algorithmic}
\usepackage{graphicx}
\usepackage{textcomp}
\usepackage{xcolor}
\def\BibTeX{{\rm B\kern-.05em{\sc i\kern-.025em b}\kern-.08em
    T\kern-.1667em\lower.7ex\hbox{E}\kern-.125emX}}

\begin{document}

% Custom commands
\newcommand{\comment}[1]{{\textcolor{red}{#1}}}
\newcommand{\rr}{\vec{r}}
\newcommand\bobold{{v1}}
\newcommand\bobnew{{v2}}

\title{Gradual Collective Upgrade of a Swarm of Autonomous Buoys for Dynamic Ocean Monitoring}

\author{\IEEEauthorblockN{Francesco Vallegra}
\IEEEauthorblockA{\textit{SUTD-MIT International Design Center} \\
\textit{Singapore University of Technology and Design}\\
Singapore
\\ francesco\_vallegra@sutd.edu.sg
}

\and

\IEEEauthorblockN{David Mateo}
\IEEEauthorblockA{\textit{Engineering Product Development} \\
\textit{Singapore University of Technology and Design}\\
Singapore
\\ david\_mateo@sutd.edu.sg
}

\and

\IEEEauthorblockN{Grgur Toki\'{c}}
\IEEEauthorblockA{\textit{Dept. of Mechanical Engineering} \\
\textit{Massachusetts Institute of Technology}\\
Cambridge, USA
\\ gtokic@mit.edu
}

\and

\IEEEauthorblockN{Roland Bouffanais}
\IEEEauthorblockA{\textit{Engineering Product Development} \\
\textit{Singapore University of Technology and Design}\\
Singapore
\\ bouffanais@sutd.edu.sg
}

\and

\IEEEauthorblockN{Dick K. P. Yue}
\IEEEauthorblockA{\textit{Dept. of Mechanical Engineering} \\
\textit{Massachusetts Institute of Technology}\\
Cambridge, USA
\\ yue@mit.edu
}

}

\maketitle

\begin{abstract}
% 1 context
Swarms of autonomous surface vehicles equipped with environmental sensors and decentralized communications bring a new wave of attractive possibilities for the monitoring of dynamic features in oceans and other waterbodies.
However, a key challenge in swarm robotics design is the efficient collective operation of heterogeneous systems.
% 2 this work
We present both theoretical analysis and field experiments on the responsiveness in dynamic area coverage of a collective of 22 autonomous buoys, where 4 units are upgraded to a new design that allows them to move 80\% faster than the rest.
% 3 results
This system is able to react on timescales of the minute to changes in areas on the order of a few thousand square meters.
We have observed that this partial upgrade of the system significantly increases its average responsiveness, without necessarily improving the spatial uniformity of the deployment.
% 4 consequences
These experiments show that the autonomous buoy designs and the cooperative control rule described in this work provide an efficient, flexible, and scalable solution for the pervasive and persistent monitoring of water environments.
\end{abstract}


\section{Introduction}
Monitoring the ocean and coastal areas has traditionally been accomplished
using either moored buoys or fixed networks of bulky, partially submerged
platforms. The cost of developing, producing, deploying, and maintaining such
systems has severely limited the range and precision of the sensed data
obtained from monitoring large-scale waterbodies. To overcome these
limitations, the community has started focusing on developing simpler,
smaller, motorized, autonomous surface
vehicles~\cite{manley2008unmanned,2009AGUSMOS22A..08O,bayat2017environmental,ziccarelli2016novel,ferreira2009autonomous}. A
large group of such vehicles operating as a collective can be deployed,
retrieved, and re-deployed (partially or completely) at considerably lower
cost and operational complexity than monolithic structures. Following the
design paradigm of swarm robotics, these collectives perform complex tasks
cooperatively in a scalable fashion, with a behavior that is both robust to
failures and flexible so as to operate in dynamic
environments~\cite{bouffanais15:_desig_contr_swarm_dynam,chamanbaz17:_swarm_enabl_techn_multi_robot_system}.
Moreover, a swarm of mobile surface vehicles forms a dynamic sensor network
suitable for high temporal sampling of waterbodies.

 % Justify heterogeneity as inevitable result of iterative development.
The pressing need for small, low-cost and rapidly deployable autonomous
vehicles has been acknowledged in multiple recent
reports~\cite{matos_man_2016,nishida_development_2015,picomultipurpose,vesecky_prototype_2007}.
However, using such large distributed systems for ocean monitoring presents as many possibilities as it does challenges. On the one hand, the system modularity and the individual cost of production primes them for iterative design and fast prototyping, in which field tests provide insights enabling a constant improvement toward  future models. On the other hand, this iterative process is likely to produce an heterogeneous system, in which  the agents have different motor, sensing, communication, or processing capabilities. This heterogeneous character must be properly handled at the system-level design in order to benefit from the range of agent capabilities.

 % Enter the BoB
 We have previously reported on the design, construction, and testing of a
 homogeneous swarm of 50 (identical) autonomous buoys performing adaptive
 deployment for applications in environmental
 monitoring~\cite{Zoss2018}. Here, we investigate and study the nontrivial process of
 partially upgrading this swarm robotics system by replacing a small fraction
 of the original swarming units (simply denoted \bobold{}) with upgraded robotic
 platforms (\bobnew{}), based on a fully
 redesigned model featuring highly improved motor and processing capabilities
 among other things.

\begin{figure}[htb]
  \centering
    \includegraphics[width=.9\linewidth]{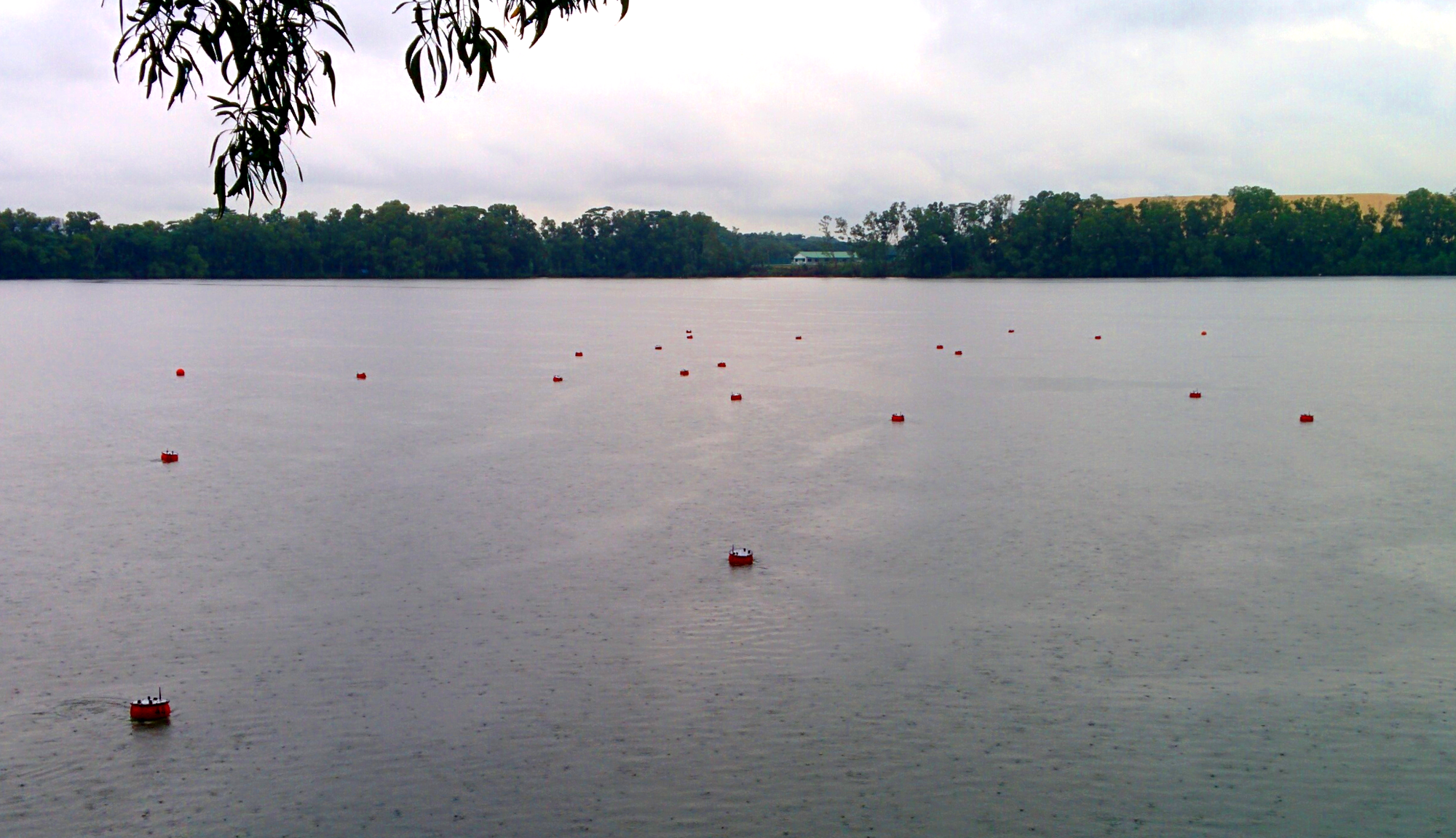}
    \caption{Swarm of buoys being deployed in the target area undergoing
      monitoring. Note that some units are still being deployed from the shore
      where the picture is taken.}
    \label{fig:photo}
\end{figure}

We have combined a set of 18 \bobold{} buoys with 4 upgraded \bobnew{} buoys---hence constituting a heterogeneous swarm---and performed dynamic area coverage experiments in fully unstructured environments without the support of any external infrastructure when it comes to the system's operations (see Fig.~\ref{fig:photo}).
We measured the collective responsiveness---a proxy to assess the flexible character of a swarm~\cite{mateo17:_effec_correl_swarm_collec_respon}---by changing the target area at different frequencies in order to investigate the minimum response time at which the heterogeneous swarm is able to adapt to changes in the monitored environment.
For this series of experiments on dynamic area coverage, the only practical differences between the two platforms are the speed of the units, their inertia, and their hull design.
Specifically, design \bobnew{} is 80\% times faster in open water and 2.3 times lighter than \bobold{}, thereby making it considerably more apt to quickly correct its position to adapt to changes in the  target area undergoing monitoring.
The field experiments performed reveal that the partial upgrade of the system is able to increase the average response of the system, but not necessarily the uniformity of the deployment.

\section{Robotic Platforms}\label{sec:fancy_buoy}

% Old Buoy
% The original buoy (\bobold{}) features a compact, omni-directional, self-righting, robust, and watertight design~\cite{Zoss2018}. A vectored propulsion system implemented with three pairs of motors allows the buoy to move through water surfaces at speeds of up to 1.0~m/s.
% The buoy is capable of self-localization using a GPS module and is designed to host a range of sensors to characterize its local environment.
% A distributed mesh communication system allows the units to exchange sensed data, send and receive commands, and broadcast their state to neighboring buoys.
% An integrated single-board computer provides the unit with enough computational power to integrate these sources of information and process them on-board in order to autonomously determine its local behavior.

In~\cite{Zoss2018}, we presented a study of dynamic ocean monitoring using a
homogeneous swarm of 50 buoys. Such decentralized and cooperative systems primarily owe their outstanding
effectiveness to the large number of agents put together: a greater number of agents allows for vaster waterbody coverage and/or refined multi-point sensing.
As technology inexorably moves forward at an increasingly faster pace, there is a compelling opportunity to expand and upgrade at least some of the agents of the collective.
These new units with improved capabilities have to be able to fit and operate within the technical framework
of the original system.

\begin{figure}
  \includegraphics[width=0.9\linewidth]{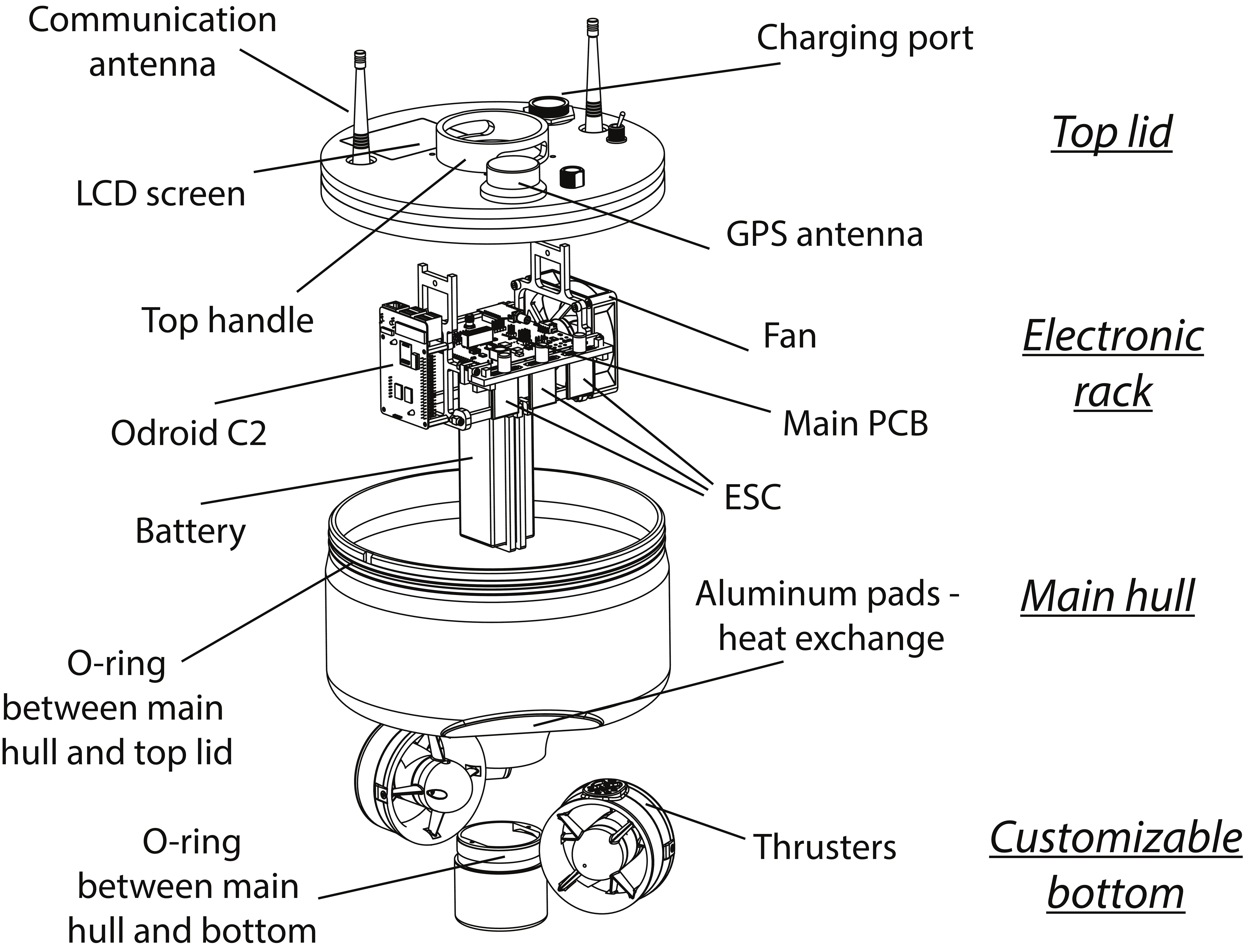}.
  \caption{Exploded view of the upgraded buoy design \bobnew{} displaying its
    modular design and main components. %\comment{ESC above stands for what?}
  }
\label{fig:explodedview}
\end{figure}

The original buoy platform---design \bobold{}---is thoroughly described
in~\cite{Zoss2018}, %where the reader can refer for full technical details.
and we refer the reader to this report for full technical details.
%about the \bobold{} design and the system design.
In this section, we focus on introducing an improved design (\bobnew{}) and its main differences with respect to \bobold{}. This new design (see Fig \ref{fig:explodedview}), re-designed mainly to facilitate operation and maintenance, can be used
%In this section, we
%focus on the description of the improved \bobnew{} design, which can be used
to replace some units in the homogeneous swarm, or expand the original homogeneous system.
%This new platform (see Fig \ref{fig:explodedview}) has been re-designed to
%facilitate operation and maintenance at scale, which is critical for such
%large-scale multi-robot systems.

% The new platform --- BoB version 2, Fig \ref{fig:explodedview} --- consists of a mechanical and electronical re-design tackling the main issues found in version 1 and focusing on a platform easier to maintain.

The platform has a modular design consisting of (i) a cylindrical hull, (ii) a
top lid hosting the electronics on one side and antennas on the other, and
(iii) a detachable bottom section that can be customized to host a number of
environmental sensors, as shown in Fig.~\ref{fig:explodedview}.
% Such modular design allows for fast inspection and faulty component replacement, aiding the maintenance process and on-field debug operations.
The main hull for \bobnew{} is designed to provide the same omnidirectional vector
propulsion apparatus as in \bobold{}, while encapsulating the three thrusters
within the cylindrical shape and keeping them as close as possible to the
center of gravity of the platform. This design choice stands in clear contrast
with \bobold{} that has protruding pairs of propellers, see
Fig.~\ref{fig:buoy_comparison}. On the one hand, this new design minimizes the
wobbling when the thrusters are active and provide a
certain level of dynamic stability to the surface vehicle.
On the other hand, the new hull has a larger packing density, which
facilitates stackability and transportation of large numbers of platforms to
remote field areas (see Fig. 1 in~\cite{Zoss2018}).
%The distribution of components inside the main hull is such that hydrodynamic
%stability is not compromised,
The mechanical design of \bobnew{} and the distribution of components
inside the main hull has been designed and tested to not compromise
the hydrostatics of \bobold{},  %and stability of \bobold{},
thereby retaining the critical self-righting feature of the \bobold{} design.

The presence of three lateral aluminum pads allows for heat exchange between
the water and the air inside the hull---a fan ensures adequate air circulation to
homogenize the heat generated
mainly by the 3 electronic speed controllers (ESC in Fig.~\ref{fig:explodedview}) and
the single-board computer Odroid C2. These aluminum pads can themselves host small sensors or light-emitting devices %for underwater communication
orientated 45 degrees below the water surface.

The top lid hosts the electronics and related hardware as a modular design
itself. An electronic rack built around the main PCB and battery is screwed on
the inside of this  lid, thereby forming one single piece that can easily be
replaced or disassembled for rapid inspection. Following the buoy \bobold{}
design, the outer part of the top lid hosts connectivity hardware such as
communication and GPS antennas, and a charging port. The top lid of \bobnew{}
includes an LCD screen reporting a range of data and indicators,
which greatly facilitates the basic inspection and troubleshooting of the
system's digital status. Lastly, a mechanical handle has been incorporated
into \bobnew{}'s top lid design to aid both manual carrying and latching a
stacked platform on top of it.

The bottom part is designed to securely host the battery, and
can be modified to accommodate any sensor.
The latter can be easily and
conveniently interfaced with the electronic stack through a mini-PCI custom
PCB that can be slotted inside the main PCB. This configuration was specifically
designed to support
further heterogeneity of the system at the sensing level, with various units
possibly carrying different sensors.

Particular attention has been placed on identifying a simple way of assembling parts and components.
An example is the thruster attachment, in which a combination of a laser-cut gaskets and screws allows for servicing all three thrusters in less than 5 minutes---eight times faster than for the \bobold{} design.
In addition, the structure of the new electronics rack is composed of laser-cut acrylic sheets and designed using screw-less joineries to facilitate manufacture, assembly and reduce potential leaks.

Once assembled, platform \bobnew{} measures $260\times260\times245$~mm and weights 3.2~kg.
For comparison, the previous design measures $350\times350\times280$~mm and weights 7.4~kg.
The reduction in dimensions is primarily achieved by moving from a spherical
form to a cylindrical one, which allows for a compact embedding of the
thrusters. The reduction in weight comes from using plastic material for the
hull as opposed to aluminum alloy in \bobold{}. Besides weight reduction, the
usage of 3D printed nylon plastic material inherently prevents corrosion and
safely withstands seawater conditions, while reducing the amount of biofouling
accumulating on the platform during operations as compared to metal.
Due to the low number of \bobnew{} prototypes, the plastic hulls have been
3D-printed, making the cost comparable with the aluminum hull of
\bobold{}. However, when mass producing the units using plastic injection
inside a mold, the cost per unit is expected to be significantly lowered.

\subsection*{Performance comparison: buoy \bobold{} vs. \bobnew{}}
\begin{figure}
\centering
  \includegraphics[width=.90\linewidth]{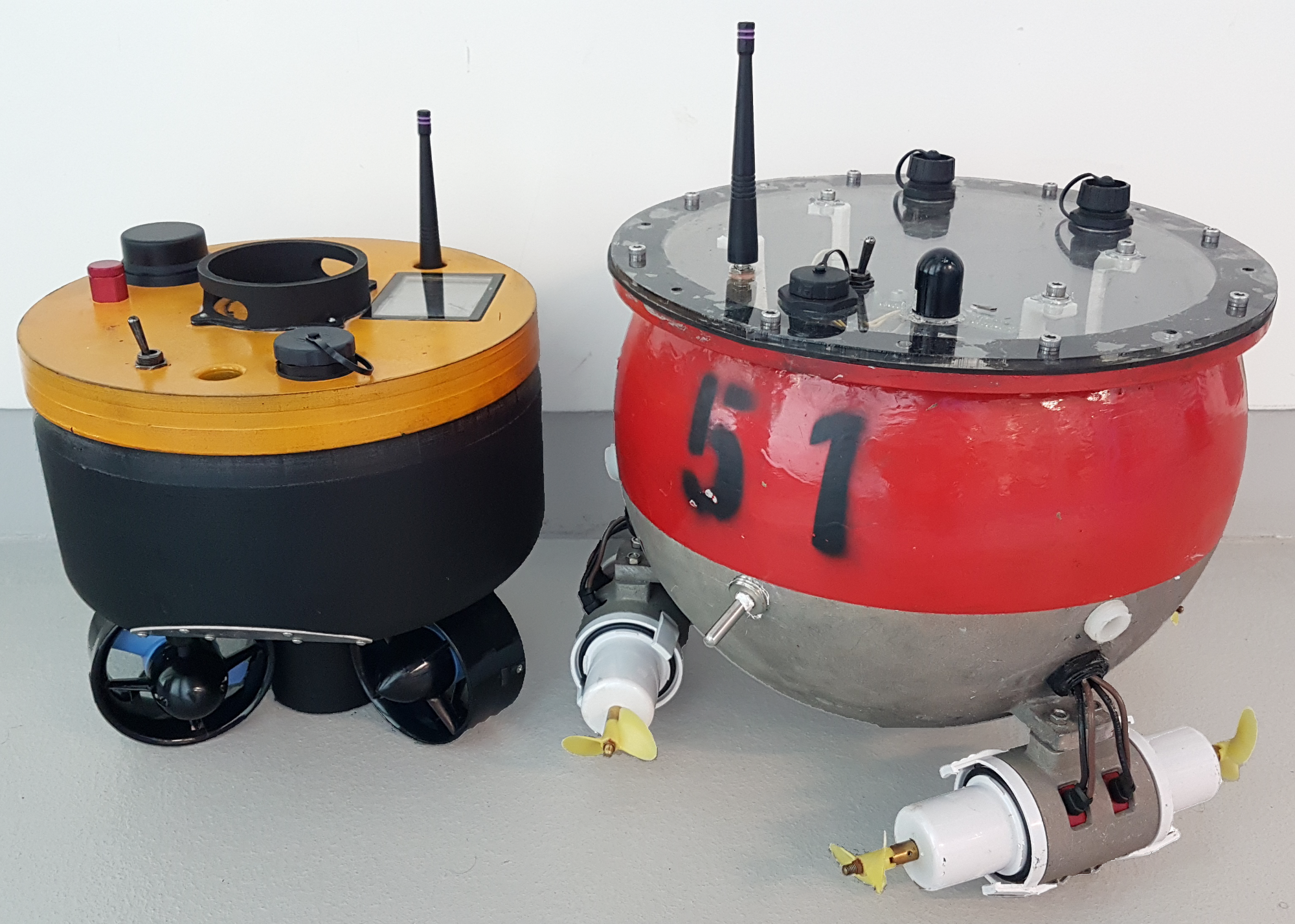}
  \caption{The two robotic platforms: the new buoy design \bobnew{} (left) side by side with one of the previous buoy model \bobnew{} (right).
  }
\label{fig:buoy_comparison}
\end{figure}

Fig.~\ref{fig:buoy_comparison} shows a visual comparison between buoy
\bobold{} and one of the prototypes of the improved design \bobnew{}. By
incorporating the thrusters into the main hull, buoy \bobnew{} is much more
compact and increases the propellers' life---the propellers are protected as
compared to \bobold{} in which the protruding pods and open propellers can
easily be damaged during transportation and/or operation.

A series of performance tests were carried out under different environmental
conditions to offer a detailed comparison of the two platforms. The tests were
run over a distance of $100$~m in the direction of the prevailing the wind and
against it. The platforms were initially positioned next to one another so as
to obtain results under the practically identical conditions. The three
BlueRobotics T100 thrusters allow the buoy \bobnew{} to move about $80$\%
faster than \bobold{} in nominal test operations, although the thrusters are
capable of performing at much higher intensity.
%This results is due to the control complexity of the platform's omnidirectional propulsion system, not yet optimized, and the overall light weight of the platform, making it move following a jerking pattern thus preventing a full thrust towards the goal. This behavior is however of little importance as the type of motion requested during operation requires fast reaction to sudden movements as opposed to long movements towards the same direction.

This self-imposed limitation on the thrusters' intensity is purely related to the need to develop a more complex vectorial propulsion system than \bobold{}'s.
However, since the cooperative control strategy remains the same for \bobold{} and \bobnew{} when performing the heterogeneous
swarming tests, it was decided to not use the full power of the
thrusters of \bobnew{}. When testing the two platforms, side by side, in
adverse environmental conditions---strong winds, waves and heavy rains, the
performance of \bobnew{} was practically unaffected. On the contrary, buoy
\bobold{} drastically underperformed, with instances at which it was fully
unable to reach its target goal.

It is worth noting that buoy \bobnew{} is granted a battery 1.5 bigger than
the one in \bobold{} to compensate the need for higher power by the thrusters,
and the extra electronics energy consumption due to fan and LCD.
%{\bf A SIMILAR POWER CONSUMPTION / DOUBLE THE POWER / \MORE? LESS?}

\section{Dynamic Area Coverage}

\subsection{Collective Operations}

Our multi-robot system can perform a range of collective behaviors
achieved by means of a cooperative control strategy supported by distributed
communications as described in full details in~\cite{Zoss2018}. For instance,
this distributed robotic system can perform a number of elementary swarming
behaviors, including flocking, navigation, and area coverage. The
effectiveness of the distributed communication setup has been verified and
analyzed in~\cite{Zoss2018}.

This large-scale networked array of mobile sensing units is designed to
monitor and characterize waterbodies, which may vary depending on the
application. %For instance, the swarm of buoys may be deployed
An example is the deployment of the swarm of buoys in a harbor to
assist in marine operations by monitoring key environmental and flow
parameters. More interestingly, the area to monitor might not be specified
externally or in advance, but instead be defined dynamically by the collective
of agents itself. By local processing of the sensed data, the agents may
determine the shape in which to self-deploy in order to track a particular
temperature profile, oil spill, or a range of biological markers.

\subsection{Cooperative Control for Dynamic Area Coverage}

As discussed in the previous section, the distributed robotic systems are designed to monitor features of interest in waterbodies (e.g. oil spills,
algal blooms, etc.). To do so, we require the agents to spread as uniformly as
possible across a given area that dynamically evolves over time.

It is paramount for this spreading to happen in a timely and responsive
way, as the shape of the area of interest is in general time-changing with
arbitrary dynamics. For this reason, we choose to define the behavior of the
agents in a purely Markovian fashion with agent's dynamical rules that
determine the buoys' movements based solely on the instantaneous state,
i.e. the cooperative control algorithm for agent $i$ can be cast as
\begin{equation}
  \frac{d\rr_i}{dt} = \vec{F}\left(t, \rr_i(t), \{\rr_j(t)\}_{j \sim i}\right) .
  \label{eq:general-control}
\end{equation}
where $\rr_i$ is the position of a given agent $i$, and $j\sim i$ is the set
of agents in the neighborhood of $i$---its ``neighbors'' according to a
specified interaction distance, be it metric, topological or else~\cite{shang14:_consen}.

The area of interest is typically determined either by external sources
(e.g. a human operator) or by the sensing capabilities of the agents
(e.g. collective tracking of a temperature gradient, or chemical concentration,
etc.). For the sake of generality, we assume that the area to monitor can be
described mathematically by
\begin{equation}
  A(\rr) < 0 ,
  \label{eq:area}
\end{equation}
where $A$ is a signed distance function (or at least a function that increases
monotonically outside the region).
Given~\eqref{eq:area}, the cooperative control rule is defined as
\begin{equation}
\frac{d\rr_i}{dt} = v_{0i} \frac{\vec{T}}{\max(1, \|\vec{T}\|)},
\end{equation}
where $v_{0i}$ is the maximum speed of an agent $i$ and $\vec{T}$ is the
``area coverage target'' defined by
\begin{equation}
\vec{T} = \frac{1}{1+\exp{(-\beta A(\rr_i))}}\frac{-\vec{\nabla}A}{\|\vec{\nabla}A\|} - \sum_{j \sim i} \frac{a_R^d}{r_{ij}^d} \frac{\rr_{ij}}{r_{ij}} \, .
\label{eq:geofencing}
\end{equation}
The first term in~\eqref{eq:geofencing}, proportional to $-\vec{\nabla}A$, attracts the agents towards the interior of the area and is scaled in such a way that its norm goes from being practically zero outside the area to cover ($A>0$) to being unity inside it ($A<0$).
The parameter $\beta$ controls how abruptly the transition between zero and
one is, i.e. the steepness of the exponential decrease.
The second term is an inter-agent repulsion term that causes the agents to spread inside the area.
The type of repulsion is controlled by two parameters, namely the repulsion strength $a_R$ and the power of repulsion $d$.
If the power $d$ is large ($d \gtrsim 4$), the repulsion strength is
approximately equal to the nearest-neighbor distance in equilibrium configurations.
% When using this ``virtual forces'' approach to maximize area coverage, one typically sets very large values like $d=32$ (ask Nikolaj for reference).
% However, one must keep in mind that large values of $d$ make the algorithm considerably more unstable and that, in realistic experimental situations, the buoys have a limited temporal and spatial resolution of their position and their neighbors'.
% This means that numerically unstable control algorithm can easily give rise to erratic behavior.

Following numerical optimization run over simulations of the behavior \eqref{eq:geofencing}, we set the free parameters to
\begin{align}
d = 6, &&  a_R = 0.38 \sqrt{S / N}, && \beta = 40 / S,
% d = 6, &&  a_R = 0.38 \sqrt{S / N}, && T = 0.025 S,
% d = 6, &&  a_R = 0.68 R_0 / \sqrt{N}, && T = 0.08 R_0^2,
\end{align}
where $S$ is the total surface area to cover and $N$ the number of agents.

\subsection{Target Surface to Monitor}
\begin{figure}
  \includegraphics{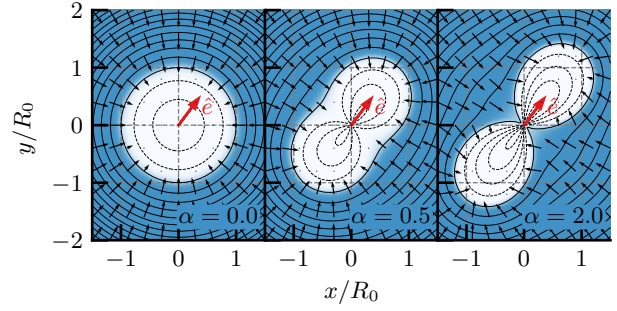}
  \caption{Effect of the geofencing term for the area given by~\eqref{eq:peanut}.
  The arrows show the vector field of the first term in~\eqref{eq:geofencing},
  and the color gradient maps its norm.
  The contour lines mark isolines $A_{\alpha,\hat{e}}$.
  }
\label{fig:forcefield}
\end{figure}

As a testbed for dynamic area exploration, we consider the following surface
\begin{equation}
  A_{\alpha,\hat{e}}(\vec{r}) = r^2 - R_{\alpha,\hat{e}}^2(\hat{r}) ,
  \label{eq:peanut}
\end{equation}
where
\begin{equation}
  R_{\alpha,\hat{e}}(\hat{r}) = \frac{R_0}{2}\frac{2 - \alpha + 3\alpha (\hat{r}\cdot\hat{e})^2}{\sqrt{1 + \alpha/2 + 11\alpha^2/32}} .
  \label{eq:radius}
\end{equation}
By changing $\alpha$, the shape of this region goes from a disk of radius $R_0$ ($\alpha=0$), to a two-lobbed area along the principal axis $\hat{e}$ ($\alpha=1$), to a ``dumbbell-shaped'' region with a nodal point at the origin ($\alpha=2$).
The normalization factor in~\eqref{eq:radius} is introduced so that the total surface of the area is kept constant at $S=\pi R_0^2$ for any $\alpha \in [0,2]$.
The gradient of this surface is
\begin{equation}
  \nabla A_{\alpha,\hat{e}}(\vec{r}) = 2 \vec{r} - 6\alpha R_0 \frac{R_{\alpha,\hat{e}}(\hat{r})}{r}(\hat{r}\cdot\hat{e})\hat{e}_\bot \, ,
\end{equation}
where $\hat{e}_\bot = \hat{e} - (\hat{r}\cdot\hat{e})\hat{r}$ is the orthogonal projection of the principal axis $\hat{e}$ on the position $\vec{r}$.
The attractive geofencing force field corresponding to this target surface is presented in Fig.~\ref{fig:forcefield}.
The contours depict the lines of constant $A_{\alpha, \hat{e}}$ and the arrows
show the vector field of the first term of~\eqref{eq:geofencing} for three values of $\alpha$.

\section{Theoretical analysis}
\label{sec:theory}

We can study the performance of the cooperative rule~\eqref{eq:geofencing} in
\emph{dynamic} area coverage by imposing a cyclic temporal evolution of the shape of the monitoring area at different frequencies.

To quantify the response of the system, we consider two metrics for the
dynamic coverage performance that were previously used when testing with a
homogeneous swarm~\cite{Zoss2018}.
The first one is the ``tessellation performance'' $P_T$, defined as the inverse of the relative size of the largest cell assigned to any agent after segmenting the target area with a Voronoi tessellation, or
\begin{equation}
    P_T = \frac{A}{A_{LVC} N},
\end{equation}
where $A$ is the total area of the target surface, $A_{LVC}$ is the area of the largest Voronoi cell (dark cell in the inset of the top panel in Fig.~\ref{fig:homo_ideal}), and $N$ is the number of agents.

The second metric is the so-called ``coverage performance'' $P_C$, defined as the percentage
of the target area covered by the agents (assuming each agent covers a disk of a certain radius $R_s$ around its position), or
\begin{equation}
    P_C = \frac{\bigcup_{i=1}^{N} {c_i} \cap A}{A},
\end{equation}
where $c_i$ is a disk of radius $R_s$ centered at the position of agent $i$
(see inset at the bottom of Fig.~\ref{fig:homo_ideal}).

The tessellation performance is meant to capture the expected accuracy of the
least accurate mobile sensing unit, since it only involves the largest Voronoi cell.
The coverage performance, on the other hand, is a measure of the average
quality of the deployment, i.e. a good proxy for system-level performance.
Both metrics take only positive values lower than unity, and where the unity
corresponds to an ideal coverage.

\subsection{Ideal Homogeneous Swarm}

We first consider a swarm of $N=20$ agents following the dynamics given by
Eq.~\eqref{eq:geofencing} with a certain speed $v_0$ identical for all the
agents. We then study the responsiveness---a proxy to assess the flexible
 character of a swarm~\cite{mateo17:_effec_correl_swarm_collec_respon}---of
 the system when covering the surface~\eqref{eq:area} changing at a frequency
 $\omega$ such that $\alpha(t) = 1 - \cos\omega t$.
The particular values of $R_0$ and $v_0$ are arbitrary, as the behavior of the system only depends on the normalized frequency $\bar{\omega} = \omega R_0 / v_0$.

% While the performance metrics $P_T$ and $P_C$ can take values between 0 and 1, in practice we find that these metrics vary within a smaller range of values such that $P_X \in [P_{X,0}, P_{X,\infty}]$ with $P_{X,0} = \lim_{\omega\rightarrow 0} P_X(\omega)$ and $P_{X,\infty} = \lim_{\omega\rightarrow \infty} P_X(\omega)$. These values are

\begin{figure}[h!]
\centering
\includegraphics{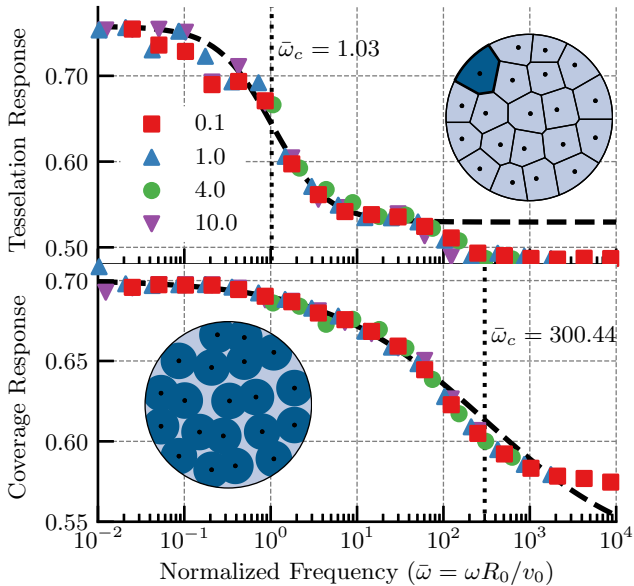}
\caption{Frequency response of a swarm of $N=20$ agents in dynamic area coverage according to its Tessellation performance $P_T$ (top) and its Coverage performance $P_C$(bottom).
The different markers correspond to calculations with different agent speeds $v_0$.
The dashed line corresponds to the ideal performance of (\ref{eq:ideal_performance}).
}\label{fig:homo_ideal}
\end{figure}

The collective response of a system following~\eqref{eq:geofencing} is shown in Fig.~\ref{fig:homo_ideal}.
The obtained response, measured either with the tessellation or coverage metrics, can be fit to the form
\begin{equation}
    P(\bar{\omega}) = \frac{P_0\bar{\omega}_c^\lambda + P_\infty\bar{\omega}^\lambda}{\bar{\omega}_c^\lambda+\bar{\omega}^\lambda},
    \label{eq:ideal_performance}
\end{equation}
where $P_0$ and $P_\infty$ are the limit performances for $\omega\rightarrow0$ and $\omega\rightarrow\infty$ respectively, $\bar{\omega}_c$ is the ``cutoff frequency'' of the system, and the exponent $\lambda$ measures how steep the transition from $P_0$ to $P_\infty$ is.
The values for these parameters obtained by fitting \eqref{eq:ideal_performance} to the results in Fig.~\ref{fig:homo_ideal} are given in Table~\ref{tab:params}.

\begin{table}[htbp]
\caption{Parameters of the ideal performance~\eqref{eq:ideal_performance} obtained by fitting the theoretical response of an ideal homogeneous system, measured both as $P_T$ and $P_C$.}
\begin{center}
\begin{tabular}{rll}
                            & Tessellation ($P_T$)  & Coverage  ($P_C$)\\
        \hline
        $P_0$               & $0.758$       & $0.700$ \\
        $P_\infty$          & $0.530$       & $0.526$ \\
        $\bar{\omega}_c$    & $1.03$        & $300.44$ \\
        $\lambda$           & $1.37$        & $0.48$ \\
\end{tabular}
\label{tab:params}
\end{center}
\end{table}

The cutoff frequency for tessellation performance is $\omega_c \simeq v_0 / R_0$.
This is revealing that the capacity of the system to maintain a uniform configuration goes down as the boundaries of the target area move too fast for a single agent to follow them.
In contrast, the cutoff frequency for the coverage performance is two orders
of magnitude larger, $\omega_c \simeq 300 v_0 / R_0$, thus showing that the
collective maintains the capacity to cover an area well beyond the individual
agents' limitations, an expected and sought feature of swarming systems.

\subsection{Ideal Heterogeneous Swarm}

\begin{figure}[h!]
\centering
\includegraphics{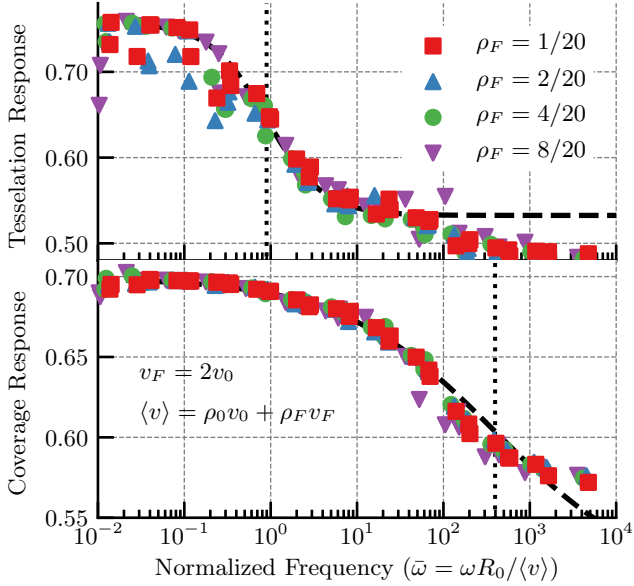}
\caption{Frequency response of a heterogeneous swarm of $N=20$ agents where a fraction $\rho_F$ of them move twice as fast as the rest, measured by its Tessellation performance (top) and Coverage performance (bottom).
The different markers correspond to calculations with different fractions of fast agents.
Note that the frequency is normalized using the average speed, which depends on $\rho_F$.}\label{fig:hetero_ideal}
\end{figure}

Next, we consider the case where the agents are not identical, and a fraction $\rho_F$ of agents move at speed $v_F = 2v_0$ while the rest move at $v_0$.
The performance of the system could in principle have a complicated dependency on $\rho_F$, $v_0$, and $v_F$.
However, we observe that as long as the two speeds differ in approximately less than one order of magnitude, the performance only depends on these parameters through the mean speed of the collective,
\begin{equation}
    \langle v \rangle = (1-\rho_F)v_0 + \rho_F v_F .
\end{equation}
The results are identical to the homogeneous case if one makes the substitution $v_0 \rightarrow \langle v \rangle$, see Fig.~\ref{fig:hetero_ideal}.
Therefore, a heterogeneous swarm with a fraction $\rho_F$ of agents moving twice as fast as the rest will have a cutoff frequency of the form
\begin{equation}
    \omega_c(\rho_F) = (1+\rho_F) \omega_c(\rho_F=0) \, .
\end{equation}

\section{Experimental Results}
\begin{figure}
    \includegraphics[width=\linewidth]{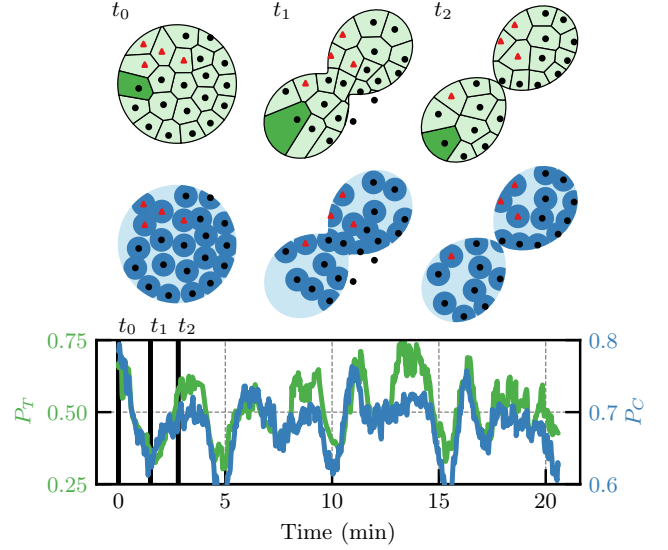}
    \caption{Evolution of the two metrics for collective response $P_T$ and $P_C$ during a field experiment. The top panels show the distribution of the 18 \bobold{} buoys (black dots) and 4 \bobnew{} ones (red triangles), along with the post-processed tessellation (top) and coverage (bottom).
    The frequency of the target area oscillations is $\omega = 0.02$~s$^{-1}$, close to the theoretically predicted cutoff frequency of the collective.}
    \label{fig:experiment_evolution}
\end{figure}

We have performed a series of field tests of dynamic area coverage using
$N=22$ buoys deployed in an uncontrolled environment with no supporting
infrastructure (Bedok Reservoir in Singapore).
For each test, the collective is tasked with covering the dynamic area defined by~\eqref{eq:area} with an $\alpha(t)$ oscillating at a certain frequency $\omega$.
Given the parameters of the test ($R_0 = 25$~m and $v_0 \simeq 0.5$~m/s), the system is expected to have a collective cutoff frequency of approximately $\omega_c = 0.02~s^{-1}$.
This means that, in general, one expects the swarm of buoys to be able to respond to changes in coverage at a time scale of approximately 1 minute when covering areas of the order of $1000~m^2$.
The evolution of the Coverage performance $P_C$, and Tessellation performance $P_T$ for one of these experiments is presented in Fig.~\ref{fig:experiment_evolution}.

\begin{figure}
    \includegraphics{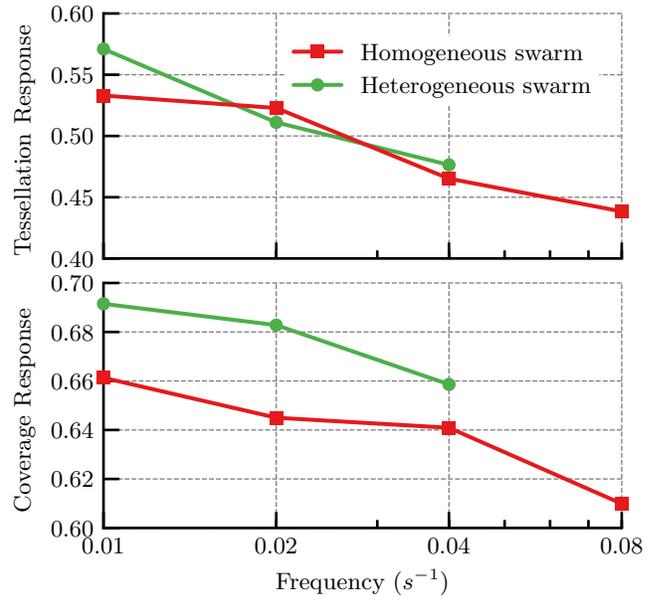}
    \caption{Experimental frequency response of a swarm of buoys tasked with dynamic area coverage.
    The homogeneous system is composed of 22 \bobold{} buoys while the heterogeneous case has 18 \bobold{} and 4 \bobnew{}.}
    \label{fig:experiment}
\end{figure}

As the core of this study is about heterogeneous swarming, we have also
carried out a series of tests to study the effect of replacing a small portion
($\rho_F=4/22$) of the $\bobold{}$ buoys by the faster and improved
model---design \bobnew{} presented in Sec.~\ref{sec:fancy_buoy}.
The theoretical prediction (see Fig.~\ref{fig:hetero_ideal}) using an
idealized model is that this heterogeneous system would display a similar
performance to the homogeneous case but with the frequencies shifted by
$\omega \rightarrow (1+\rho_F)\omega = 1.18 \omega$.
The experimental results for a total of 7 field tests are presented in Fig.~\ref{fig:experiment}.
The overall performance of the homogeneous system, measured either by the
tessellation or coverage metrics, is about $25\%$ lower than the ideal case.
This is a consequence of {\it (i)} the real dynamics and controllability of
the buoys, {\it (ii)} the precision and accuracy of the GPS-based
localization, and {\it (iii)} the finite communication rate between buoys as
part of the distributed communication network.
All these factors contribute to some extent, to the fact that the spatial
distribution of the buoys is not as uniform as the cooperative control algorithm allows for.

Introducing heterogeneity in the real swarming system yields interesting
results regarding its collective response (see
Fig.~\ref{fig:experiment}).
Instead of slightly improving both performance
metrics as in the ideal case (see Sec.~\ref{sec:theory}), what we observe
is that a small proportion ($\rho_F=4/22\simeq 18 \%$) of faster buoys
dramatically improves the coverage performance of the system while having a
negligible effect on the tessellation performance.

As discussed before, the tessellation metric measures the performance of the
system at the location where the deployment is the {\it least} uniform, and it
is therefore primarily sensitive to the worst-performing section of the system.
The coverage metric, in contrast, takes into account the deployment of all
buoys and thus measures the system-level performance of this swarming system.
These experimental observations reveal how the heterogeneity affects the operations of the system.
Since the agents form a regular lattice in the target area and have a relatively small mobility within that lattice, the faster buoys (\bobnew{}) can only use their higher speed to improve the deployment in the vicinity of their location in the lattice.
This improvement always translates into an improved average performance, and thus an improved coverage.
However, if the worst-performing agent happens to be far from the \bobnew{} buoys---e.g. in the snapshots of Fig.~\ref{fig:experiment_evolution}, these cannot affect its performance, and thus the tessellation performance remains virtually the same as in the heterogeneous case.

\section{Discussion}

The efficient collective operation of heterogeneous swarms has been identified as one of the key challenges in robotics~\cite{Yangeaar7650}.
In general, such a swarm can be composed of a large number of different
classes of agents where some may be specialized in specific tasks such as
sensing, communications, or (sensed) data processing.
However, with complex systems yielding emergent behaviors, it is not obvious
that such individual enhancements of a fraction of the agents automatically translate into an increased collective performance.
In this work, we study the effect of increasing the motor capabilities of
18\% of the agents on the responsiveness of the group when performing a dynamic area monitoring in real-world and unstructured water environments.

Experimentally, we observe that the timescale at which the system is able to respond is in good agreement with the theoretical prediction, confirming that this distributed smart sensor array of buoys is capable of responding to changing environments on the order of the minute, which is well beyond what is needed to track morphological changes in oil spills, algal blooms, or other surface contaminants.

We also present empirical evidence that the partial upgrade of the system
improves the responsiveness of the system when using a cooperative control
algorithm designed for homogeneous systems and that does not explicitly take
into account the different motor capabilities of the heterogeneous set of agents.
We measure the performance of the system with two metrics, the tessellation $P_T$ and the coverage $P_C$ performances.
Since $P_T$ only takes into account the agent with the largest area to cover, it is an individual measure of the ``weakest link'' in the system.
As such, it is indicative of the robustness of the collective.
$P_C$, on the other hand, measures an average collective performance that is less sensitive to any individual agent's behavior.
While the partial upgrade makes the system more capable of dynamically covering a target area on average ($P_C$), it does not improve the coverage of the least-covered regions ($P_T$).

In order to magnify the effect of the new and improved agents, the system
would need to operate taking into account the specific properties of the
different agents. For instance, a further improvement in the collective
responsiveness of this heterogeneous swarm could be achieved by developing a new cooperative control strategy
that positions the faster agents in the most critical areas, or that
implements an optimization procedure constrained by the motor capabilities of
the different agents.
Such a procedure, however, could prove difficult to maintain as it is not robust to hardware design changes: as more kinds of agents are added during the life-cycle of the collective, the control algorithm should grow in complexity to accommodate these changes.

\section*{Acknowledgments}
This work was supported by Grants from the SUTD-MIT International Design
Center (IDC) and the Singapore Ministry of Education (MOE-Tier 1 Grant \#T1MOE1701).
We are grateful for the assistance of B. Patel, H. Shekhar, S. Jain, and P. Rastogi in preparing and performing the field tests described in this work.

% \bibstyle{IEEEtran}
% \bibliography{refs}
% \bibliographystyle{ieeetr}

\end{document}